\documentclass{article}

\usepackage{amssymb}
\usepackage{amsmath}
\usepackage{amsthm}
\usepackage{graphicx}
\usepackage{subfigure}

\usepackage{nips14submit_e,times}

\nipsfinalcopy % Uncomment for camera-ready version

\begin{document}

\title{Some Theory For Practical Classifier Validation}

\author{Eric Bax\thanks{baxhome@yahoo.com} \hbox{ } and \hbox{ }Ya Le\thanks{yle@stanford.edu}}

%\editor{??}

\maketitle

\begin{abstract}
We compare and contrast two approaches to validating a trained classifier while using all in-sample data for training. One is simultaneous validation over an organized set of hypotheses (SVOOSH), the well-known method that began with VC theory. The other is withhold and gap (WAG). WAG withholds a validation set, trains a holdout classifier on the remaining data, uses the validation data to validate that classifier, then adds the rate of disagreement between the holdout classifier and one trained using all in-sample data, which is an upper bound on the difference in error rates. We show that complex hypothesis classes and limited training data can make WAG a favorable alternative. 
\end{abstract}

\section{Introduction}
One goal in machine learning is to use all available data for training and still compute effective test error bounds. A seminal result by Vapnik and Chervonenkis \cite{vapnik71} showed that we can use the same data for training and validation, if we use simultaneous validation over all hypotheses in the hypothesis class used for training. Using this setup, the probability that test error rate for the trained classifier exceeds its training error rate by at least $\epsilon$ is at most
$$ m(n) b(n,\epsilon), $$
where $n$ is the number of training examples, $m(n)$ is the number of hypotheses in the hypothesis class, and $b(n,\epsilon)$ is an upper bound on a deviation of at least $\epsilon$ between empirical and actual means over $n$ samples. This bound was revolutionary: since $m(n)$ grows as a polynomial in $n$ for many hypothesis classes and $b(n,\epsilon)$ shrinks exponentially in $n$, we can be confident that if training selects a classifier with a low training error rate, it will have a low test error rate as well, with high probability, given sufficiently many training examples.

The original VC bound used shattered sets and VC dimension to bound $m$. There have been many improvements and variations on hypothesis counting, including the luckiness framework \cite{cristianini00}, margin bounds for hyperplanes\cite{vapnik98}, and PAC-Bayesian bounds \cite{mcallester99}. These advances all build on the same basic concept of validation based on the number of hypotheses in a hypothesis class (or a distribution over those hypotheses), selected prior to examining the training data. 

Similarly, there have been many improvements and variations on the Hoeffding bounds \cite{hoeffding63} originally used to bound $b$, including Azuma bounds \cite{azuma67}, McDiarmid bounds \cite{mcdiarmid89}, direct computation of bounds by binomial inversion \cite{langford05,hoel54}, bounds that take advantage of the low variance for accurate classifiers \cite{audibert07,maurer09}, and other concentration inequalities \cite{boucheron13}. These results focus on $b$, so they can be applied to a single classifier as easily as to a bound over multiple hypotheses. 

Consider an alternative validation method: withold $v < n$ validation examples. Train a \textit{holdout classifier} on the remaining $n-v$ examples. Next, train a \textit{full-data classifier} on all training examples. Then compute (in the transductive setting, where test inputs are known \cite{vapnik98}) or validate the rate of disagreement $\Delta$ between the holdout and full-data classifiers over test data. 

Since the holdout classifier is independent of the validation examples, we can use single-classifier validation over those examples to bound the test error rate of the holdout classifier. The full-data classifier test error rate is at most $d$ greater than the holdout classifier test error rate. (In the worst case, every disagreement is an error for the full-data classifier.) So the probability that the full-data classifier test error rate exceeds the holdout classifier error rate over the validation data by at least $\epsilon$ is at most 
$$ b(v, \epsilon - \Delta) $$
in the transductive case. (We will concentrate on the transductive case in the main body of this note.) Call this the WAG (withhold and gap) bound.

Compare the WAG bound to the bound based on validation over a set of hypotheses (SVOOSH). The WAG bound has the advantage of not requiring simultaneous bounds over $m$ hypotheses. However, it uses fewer examples for validation ($v$ vs. $n$), and it has to include the rate of disagreement $\Delta$ as part of the bound range, $\epsilon$. 

In this note, we explore conditions for WAG to provide stronger error bounds than traditional methods. We also offer some intuition around the process of validation, considering for the traditional methods how many examples are needed to ``select a hypothesis'' vs. to ``validate the selected hypothesis" and for WAG how similarity between a holdout and the full-data classifier can indicate that learning has been effective.

\section{The Cost of Simultaneous Validations}
For SVOOSH, consider how many extra examples are needed because we validate multiple hypotheses instead of just one. For SVOOSH, 
$$ \delta = m(n) b(n,\epsilon). $$
Using Hoeffding bounds \cite{hoeffding63}, 
$$ b(n,\epsilon) = e^{-2n\epsilon^2}. $$
So
$$ \delta = m(n) e^{-2n\epsilon^2}. $$
Let $s$ be the value such that 
$$ e^{-2s\epsilon^2} = \frac{1}{m(n)}. $$ \label{smile}
Then
$$ \delta = m(n) e^{-2s\epsilon^2} e^{-2(n-s)\epsilon^2} = e^{-2(n-s)\epsilon^2} = b(n-s, \epsilon), $$ \label{oh}
which is the probability probability of bound failure for single-classifer validation using $n-s$ examples. 

Solve Equality \ref{smile} for $s$:
$$ s = \frac{\ln m(n)}{2 \epsilon^2}. $$ \label{argh}
Since $s$ is O($\ln m$), the number of extra examples required to validate multiple hypotheses grows very slowly in the number of hypotheses. Suppose $m(n) = n^d$, and call $d$ the \textit{dimension} of the hypothesis set -- similar to VC dimension. Then $s$ is O($d \ln n$). So $s$ depends strongly on dimension $d$. Also, the portion of examples required because of multiple-hypothesis validation, $\frac{s}{n}$, shrinks quickly with the number of examples: it is O($\frac{\ln n}{n}$). 

\section{A Limit on Validation Set Size for WAG}
What does this have to do with WAG vs. SVOOSH? In SVOOSH, $\delta = b(n-s, \epsilon)$, meaning that the bound is the same as for using $s$ examples to select a hypothesis from the set and then using the remaining $n-s$ examples to validate the selected hypothesis. In WAG, with $\delta = b(v, \epsilon - \Delta)$, we use $n-v$ examples to select a hypothesis, then use the remaining $v$ examples to validate that hypothesis. For WAG to offer a stronger bound than SVOOSH, we must have $b(v, \epsilon - \Delta) < b(n-s, \epsilon)$. (For concentration inequalities $b()$ other than Hoeffding's inequality, there exists a minimum $s$ such that $b(n-s, \epsilon) \leq m(n) b(n, \epsilon)$, and that $s$ value plays the same role as our $s$. In luckiness frameworks and PAC-Bayesian frameworks, the $s$ may depend on the selected hypothesis.)

The best case for WAG is $\Delta = 0$ -- no disagreement between the holdout classifier and the one trained on all data. Even for this case, $b(v, \epsilon - \Delta) < b(n-s, \epsilon)$ requires $v > n-s$. So $n-s$ is a lower bound on the validation set size $v$ if WAG is to be superior to SVOOSH. From Equality \ref{argh}, 
$$ n-s = n - \frac{d}{2 \epsilon^2} \ln n. $$
So
$$ v > n - \frac{d}{2 \epsilon^2} \ln n $$
is required for WAG to be superior. That leaves at most 
$$
w^* = \frac{d}{2 \epsilon^2} \ln n $$
examples for training the holdout classifier in WAG. 

In general, using fewer examples to train the holdout classifier results in a greater rate of disagreement $\Delta$. So, for WAG to be more effective than SVOOSH, we can see that the hypothesis set dimension $d$ must be large or the number of training examples $n$ must be small, since $\ln n$ is a quickly-shrinking portion of $n$ as $n$ increases. 

\section{WAG vs. SVOOSH}
To compare WAG to SVOOSH, let us examine the values of $\Delta$ needed for WAG to outperform SVOOSH. Set bound faliure probability $\delta$ equal for both methods, let $\epsilon_W$ be the bound range for WAG, and let $\epsilon_V$ be the bound range for SVOOSH. Then
$$ \epsilon_W = \Delta + \sqrt{\frac{\ln \frac{1}{\delta}}{2 v}}, $$
and 
$$ \epsilon_V = \sqrt{\frac{\ln \frac{1}{\delta} + \ln m(n)}{2 n}}. $$
Let $m = n^d$. For $a>1$, use $v = n/a$ examples for validation. Then $\epsilon_W < \epsilon_V$ (WAG superior to SVOOSH) requires 
$$ \Delta < \frac{1}{\sqrt{2 n}} \left(\sqrt{\ln \frac{1}{\delta} + d \ln n} - \sqrt{a \ln \frac{1}{\delta}}\right). $$
Let $\Delta^*$ be the critical value for $\Delta$:
$$ \Delta^* = \frac{1}{\sqrt{2 n}} \left(\sqrt{\ln \frac{1}{\delta} + d \ln n} - \sqrt{a \ln \frac{1}{\delta}}\right). $$

\begin{figure}
  \includegraphics[width=\linewidth]{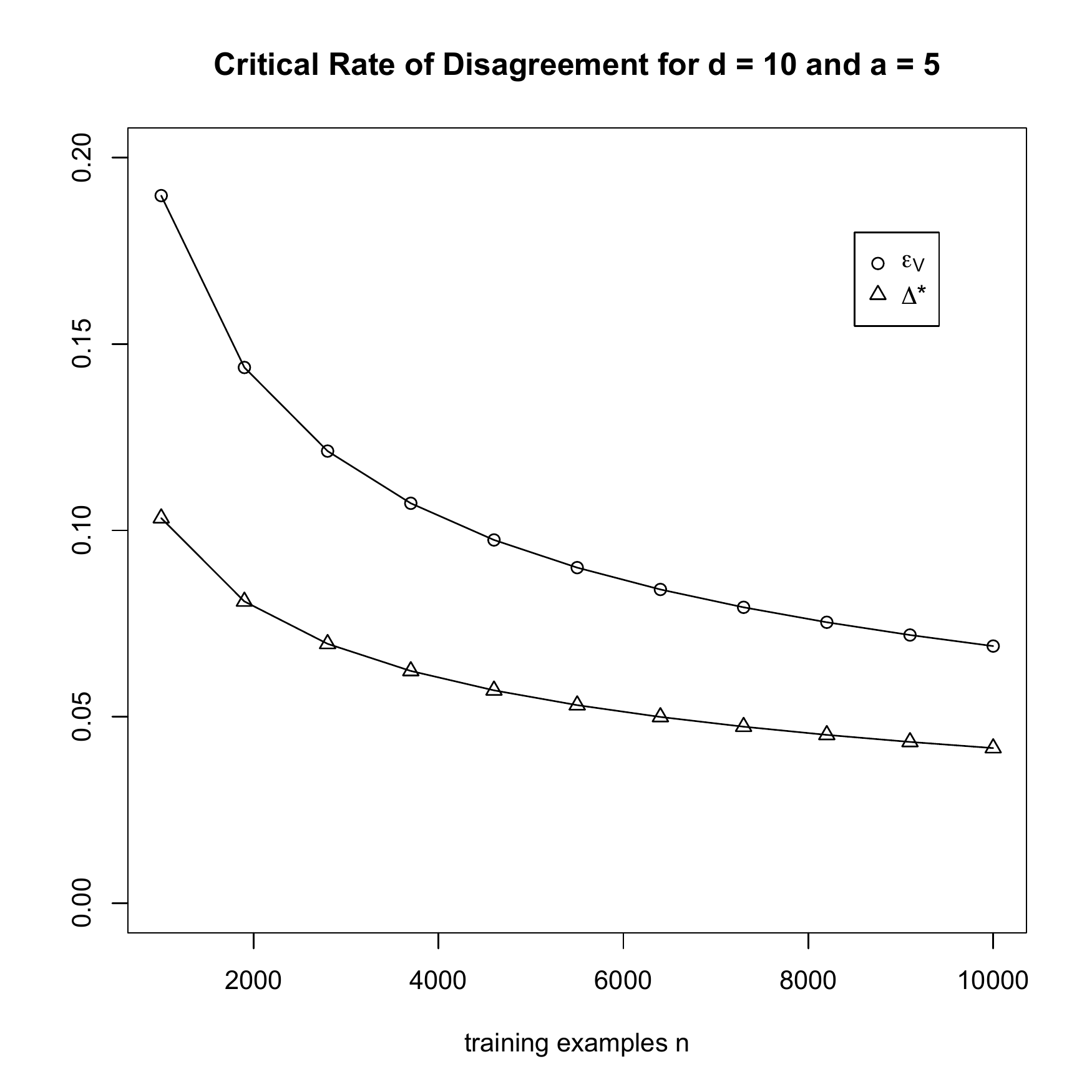}
  \caption{If there is less than about five to ten percent disagreement between holdout and full-data classifiers, WAG outperforms SVOOSH.}
  \label{fig:d10}
\end{figure}

\begin{figure}
  \includegraphics[width=\linewidth]{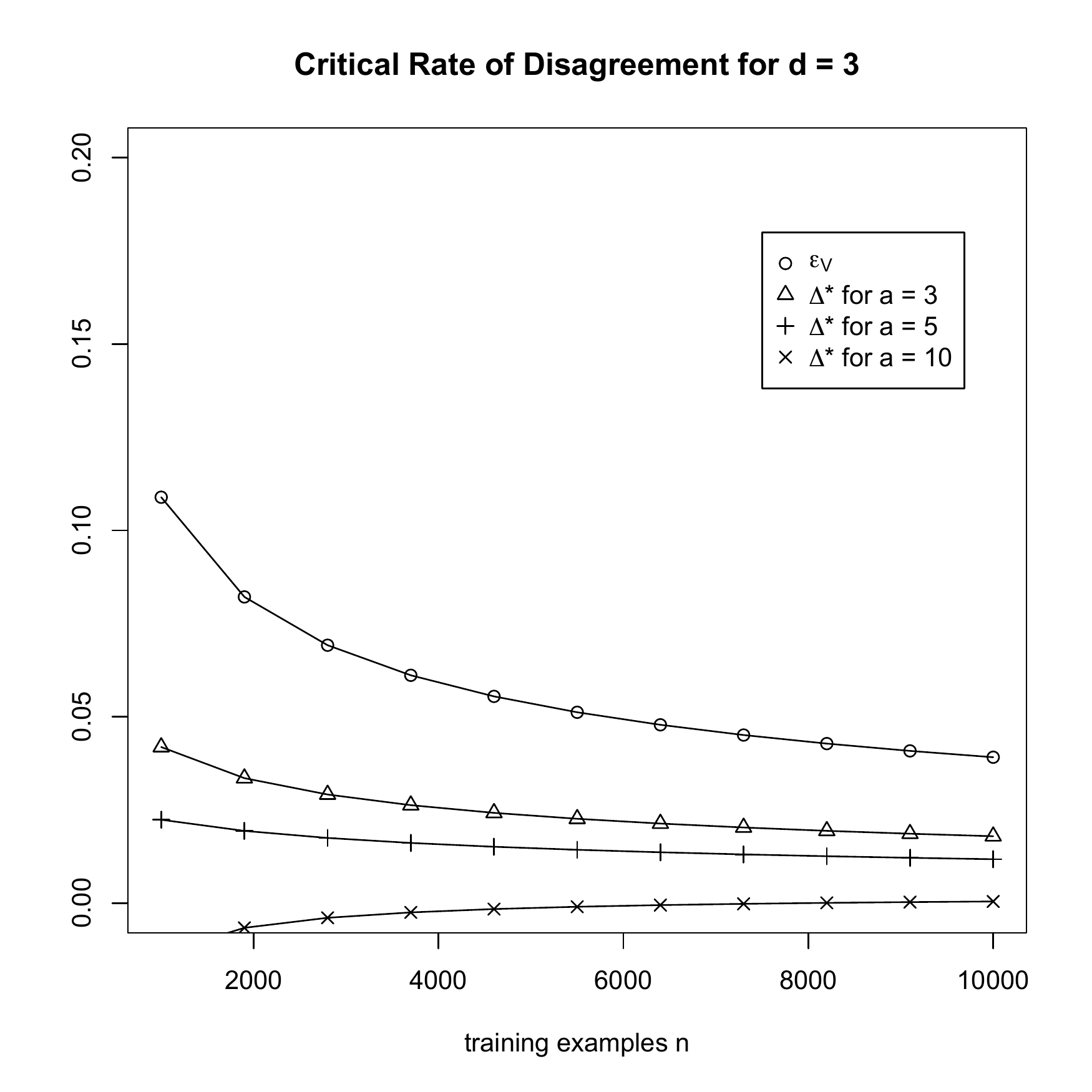}
  \caption{A lower-complexity hypothesis class leaves less room for WAG to outperform SVOOSH.}
  \label{fig:d3}
\end{figure}

\begin{figure}
  \includegraphics[width=\linewidth]{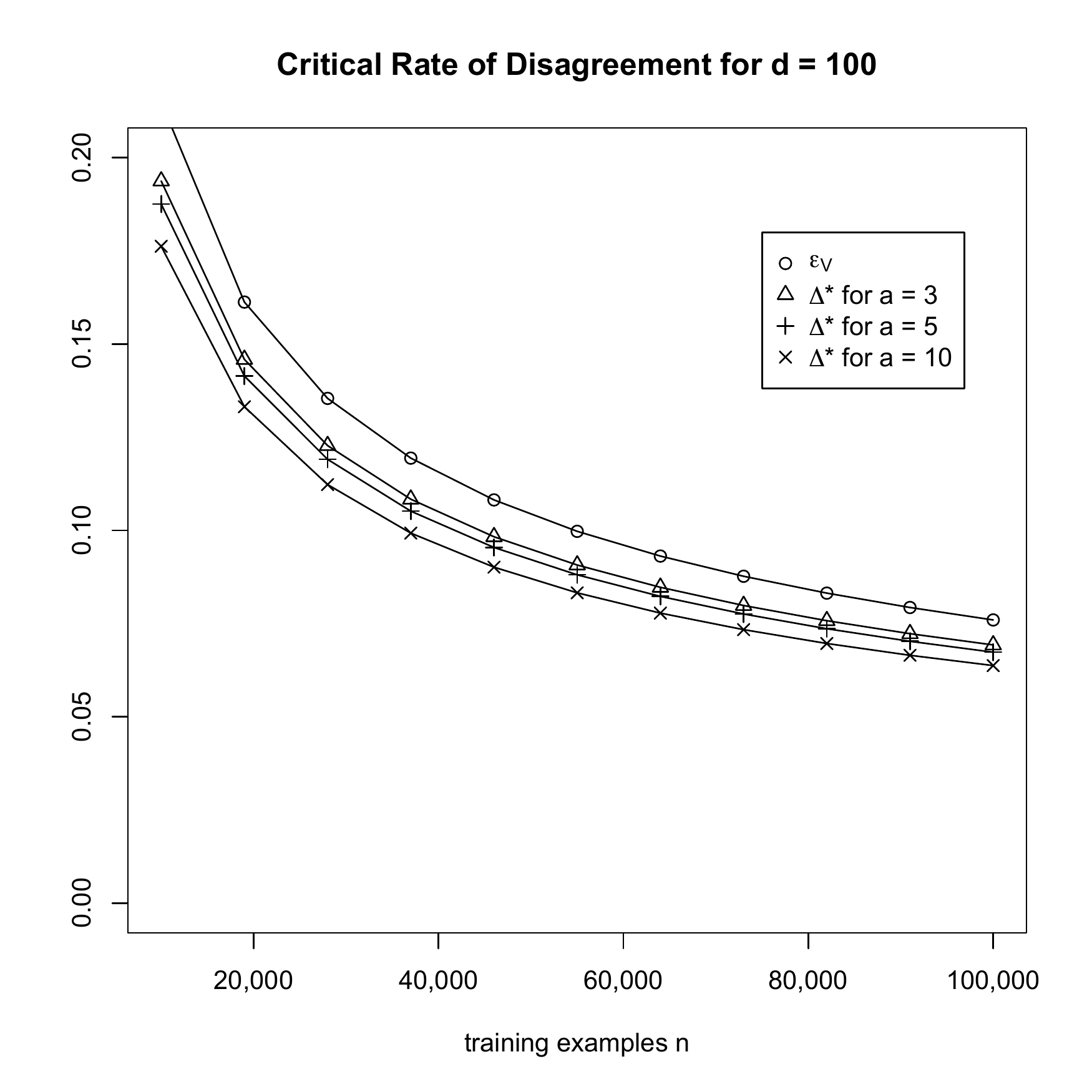}
  \caption{A higher-complexity hypothesis class makes more room for WAG to outperform SVOOSH. }
  \label{fig:d100}
\end{figure}

Figures \ref{fig:d10}, \ref{fig:d3}, and \ref{fig:d100} show some values of $\Delta^*$ and $\epsilon_V$. For all the plots, bound failure probability $\delta = 0.05$.  
Begin with Figure \ref{fig:d10}. It shows $\epsilon_V$, the bound range for SVOOSH, and $\Delta^*$, the critical rate of disagreement between holdout and full-data classifiers for WAG to outperform SVOOSH -- below this rate of disagreement, WAG outperforms SVOOSH. For this plot, $d=10$, meaning that the hypothesis class has dimension 10 ($m(n) = n^{10}$), and $a=5$, meaning that one fifth of the examples are used for validation ($v = n/5$) and four fifths are used to train the holdout classifier. From 1000 to 10,000 training examples,  $\Delta^*$ varies from about 0.10 to about 0.04, meaning that the rate of disagreement between a classifier trained on $4/5$ of the examples and one trained on all examples must be less than about five to ten percent for WAG to be superior. This does not seem unreasonable.

The single-classifier bound range $\epsilon_W$ is the difference between the plotted lines: $\epsilon_V - \Delta^*$. The single-classifier bound range is about half the multiple-classifier bound range for all the numbers of examples $n$ shown in the graph. (This is multiple-classifier with $n$ examples vs. single-classifier with $n/5$ examples.) This makes $\Delta^*$ about half the multiple-classifier bound range. So $\Delta^*$ is greatest for low numbers of training examples, where the multiple-classifier bound renege is greatest. 

Figure \ref{fig:d3} shows $\Delta^*$ and $\epsilon_V$ for a less complex hypothesis class, with $d=3$. With fewer hypotheses, the multiple-classifier bound range $\epsilon_V$ decreases compared to Figure \ref{fig:d10}. But the single-classifier bound range $\epsilon_W$ (not shown) remains the same. As a result, $\Delta^*$, which is the difference between these bound ranges, gets squeezed. The figure shows $\Delta^*$ for $a = 3$, 5, and 10, meaning that the validation sets are one third, one fifth, and one tenth of the training examples. For $a = 3$ and $a = 5$, $\Delta^*$ ranges from about one to five percent. 

For $a = 10$, there are too few validation examples to perform effective enough single-classifier validation using $n/10$ examples to leave room for a reasonable $\Delta^*$. For the lower values of $n$, the single-classifier bound based on one tenth of the examples is in fact worse than the multiple-classifier bound based on all examples. That makes $\Delta^*$ negative. 

Figure \ref{fig:d10} shows $\Delta^*$ and $\epsilon_V$ for $d = 100$. Due to the complexity of the hypothesis class, more training examples are required for SVOOSH to yield reasonable error bounds. The single-classifier bound ranges using validation data ($\epsilon_W$, the gaps between $\epsilon_V$ and $\Delta^*$) are several times smaller than the multiple-classifier bound ranges using all training data ($\epsilon_V$). That leaves room for $\Delta^*$ values ranging from about seven to nineteen percent.

\section{Discussion}
We showed that WAG can outperform SVOOSH for validation of classifier training, using reasonable validation data set sizes, when the hypothesis set is sufficiently complex and there are few enough training examples, assuming reasonable rates of disagreement between holdout and full-data classifiers. After some consideration, WAG makes intuitive sense. If adding a few more training examples radically changes the resulting classifier, then it is hard to believe that the original (holdout) classifier would have generalized well. Similarly, it is easy to believe that adding a few more examples would result in yet again a very different classifier. In that case, it is hard to believe that the full-data classifier will generalize well.

How can we measure rate of disagreement $\Delta$ in a nontransductive setting? If there are unlabeled examples that are not used for learning, then we can use them to evaluate the rate of disagreement. If these are not the test examples, then the rate of disagreement over them is an empirical mean, and we can use concentration inequalities $b(t, \epsilon_T)$, where $t$ is the number of unlabeled examples, to bound the value of $\Delta$ over their generating distribution -- presumably the same as the distribution of test examples. In this case, the WAG bound range is the sum of $\epsilon_W$ and $\epsilon_T$. Fortunately, unlabeled examples are often easier to obtain than labeled ones, so $\epsilon_T$ can often be made small by taking a large number of unlabeled samples $t$. 

It may be tempting to assert that the withheld validation examples can be used to simultaneously validate the holdout classifier and the rate of disagreement between the holdout classifier and the full-data classifier. However, the full-data classifier is not independent of the withheld data, since that data is part of its training set. An open question is whether there are conditions on training algorithms or hypothesis classes that allow this approach. 

Recently, WAG has been used to validate network classifiers \cite{bax13} and matching algorithms \cite{le14}. Central classifier bounds \cite{bax97} are similar to the WAG approach analyzed here, but apply to validation of a classifier selected by early stopping. The method includes withholding a data set for validation and using it to validate the rate of disagreement between classifiers, but not using the validation data to train a full-data classifier. The general idea of withholding a data set for evaluation and then bounding the gap between the holdout classifier and a full-data classifier was first used for nearest neighbor classifiers, which have quite fractured decision boundaries, making it difficult to assess the size of their hypothesis sets. These methods \cite{devroye79,devroye96} do not empirically evaluate the rate of disagreement between holdout and full-data classifiers; instead they take advantage of the locality of nearest neighbor classifiers and to prove upper bounds for the rates of disagreement based on geometry. As a result, they can be applied outside of transductive settings. More recent methods for nearest neighbors \cite{bax00,bax12} use empirical evaluation of rates of disagreement, including inclusion-exclusion techniques \cite{bax15}. But they still rely on the locality of nearest neighbors classifiers to evaluate the gap between holdout and full-data classifiers, so it remains an open question whether such approaches can be applied to other types of classifiers. 

\bibliographystyle{unsrt}
\bibliography{bax}

\begin{thebibliography}{10}

\bibitem{vapnik71}
V.~Vapnik and A.~Chervonenkis.
\newblock On the uniform convergence of relative frequencies of events to their
  probabilities.
\newblock {\em Theory of Probability and its Applications}, 16:264--280, 1971.

\bibitem{cristianini00}
N.~Cristianini and J.~Shawe-Taylor.
\newblock {\em An Introduction to Support Vector Machines and Other
  Kernel-Based Learning Methods}.
\newblock Cambridge University Press, 2000.

\bibitem{vapnik98}
V.~Vapnik.
\newblock {\em Statistical Learning Theory}.
\newblock John Wiley \& Sons, 1998.

\bibitem{mcallester99}
David McAllester.
\newblock Some {PAC-Bayesian} theorems.
\newblock {\em Machine Learning}, 37(3):355--363, 1999.

\bibitem{hoeffding63}
W.~Hoeffding.
\newblock Probability inequalities for sums of bounded random variables.
\newblock {\em Journal of the American Statistical Association},
  58(301):13--30, 1963.

\bibitem{azuma67}
K.~Azuma.
\newblock Weighted sums of certain dependent random variables.
\newblock {\em T™hoku Mathematical Journal}, 19(3):357Ð367, 1967.

\bibitem{mcdiarmid89}
C.~McDiarmid.
\newblock On the method of bounded differences.
\newblock {\em Suveys in Combinatorics, London Math. Soc. Lecture Notes},
  141:148--188, 1989.

\bibitem{langford05}
J.~Langford.
\newblock Tutorial on practical prediction theory for classification.
\newblock {\em Journal of Machine Learning Research}, 6:273--306, 2005.

\bibitem{hoel54}
P.~G. Hoel.
\newblock {\em Introduction to Mathematical Statistics}.
\newblock Wiley, 1954.

\bibitem{audibert07}
J.-Y. Audibert, R.~Munos, and Csaba Szepesvari.
\newblock Variance estimates and exploration function in multi-armed bandit.
\newblock {\em CERTIS Research Report 07-31}, 2007.

\bibitem{maurer09}
Andreas Maurer and Massimiliano Pontil.
\newblock Empirical {Bernstein} bounds and sample-variance penalization.
\newblock In {\em COLT}, 2009.

\bibitem{boucheron13}
S.~Boucheron, G.~Lugosi, and P.~Massart.
\newblock {\em Concentration Inequalities -- A Nonasymptotic Theory of
  Independence}.
\newblock Oxford University Press, 2013.

\bibitem{bax13}
E.~Bax, J.~Li, A.~Sonmez, and Z.~Cataltepe.
\newblock Validating collective classification using cohorts.
\newblock {\em NIPS Workshop on Frontiers of Network Analysis: Methods, Models,
  and Applications}, 2013.

\bibitem{le14}
Ya~Le, Eric Bax, Nicola Barbieri, David~Garcia Soriano, Jitesh Mehta, and James
  Li.
\newblock Validation of network reconciliation.
\newblock {\em http://arxiv.org/abs/1411.0023}, 2015.

\bibitem{bax97}
Eric Bax, Zehra Cataltepe, and Joseph Sill.
\newblock The central classifier bound Ð a new error bound for the classifier
  chosen by early stopping.
\newblock {\em IEEE PACRIM Conference, Victoria, Canada}, pages 811--814, 1997.

\bibitem{devroye79}
L.~Devroye and T.~Wagner.
\newblock Distribution-free inequalities for the deleted and holdout estimates.
\newblock {\em IEEE Transactions on Information Theory}, 25:202--207, 1979.

\bibitem{devroye96}
L.~Devroye, L.~Gy{\"o}rfi, and G.~Lugosi.
\newblock {\em A Probabilistic Theory of Pattern Recognition}.
\newblock Springer, 1996.

\bibitem{bax00}
E.~Bax.
\newblock Validation of nearest neighbor classifiers.
\newblock {\em IEEE Transactions on Information Theory}, 46(7):2746--2752,
  2000.

\bibitem{bax12}
E.~Bax.
\newblock Validation of $k$-nearest neighbor classifiers.
\newblock {\em IEEE Transactions on Information Theory}, 58(5):3225--3234,
  2012.

\bibitem{bax15}
Eric Bax, Lingjie Weng, and Xu~Tian.
\newblock Validation of k-nearest neighbor classifiers using inclusion and
  exclusion.
\newblock {\em http://arxiv.org/abs/1410.2500}, 2015.

\end{thebibliography}

\end{document}